\documentclass[journal,twoside,web]{ieeecolor}
\usepackage{generic}
\usepackage{cite}
\usepackage{amsmath,amssymb,amsfonts}
\usepackage{algorithmic}
\usepackage{graphicx}
\usepackage{textcomp}
	\usepackage{cite}
\usepackage{algorithm, algorithmic}
\usepackage{amsfonts}
\usepackage{amsopn}
\usepackage{amsmath}
\usepackage{amstext}
\usepackage{amsbsy}
\usepackage{subfigure}
\usepackage{graphicx}
\usepackage{float}
\usepackage{multirow}
\usepackage{cellspace}
\usepackage{longtable}
\usepackage{stfloats}
\usepackage{booktabs}
\usepackage{array}
\usepackage{footnote}
\usepackage{subfigure}
\makesavenoteenv{tabular}

\usepackage[colorlinks,linkcolor=blue]{hyperref}
\usepackage{indentfirst}
\usepackage{amssymb, bm}
\usepackage{ragged2e}
\usepackage{subfigure}
\usepackage{array}

\usepackage{color}
\usepackage{soul, color,xcolor}
\soulregister\ref7
\soulregister\cite7

\def\BibTeX{{\rm B\kern-.05em{\sc i\kern-.025em b}\kern-.08em
    T\kern-.1667em\lower.7ex\hbox{E}\kern-.125emX}}
\begin{document}
\title{Real-time 3D human action recognition based on Hyperpoint sequence}
\author{Xing~Li, ~\IEEEmembership{Student Member,~IEEE}, Qian~Huang, Zhijian~Wang, Tianjin~Yang, Zhenjie~Hou, and~Zhuang~Miao
\thanks{Manuscript received May 6, 2022. This work is partly supported by the National Key Research and Development Program of China under grant No. 2018YFC0407905. (Corresponding authors: Qian Huang.) }
\thanks{Xing Li, Qian Huang, Zhijian Wang, and Tianjin Yang are with the Key Laboratory of Water Big Data Technology of Ministry of Water Resources, Hohai University; and School of Computer and Information, Hohai University, Nanjing 211100, China(e-mail: huangqian@hhu.edu.cn).}
\thanks{Zhenjie Hou is with Changzhou University, Changzhou 213000, China.}
\thanks{Zhuang Miao is with Command and Control Engineering College, Army Engineering University of PLA, Nanjing 210007, China(e-mail: emiao$\_$beyond@163.com).}
}

\maketitle

\begin{abstract}
Real-time 3D human action recognition has broad industrial applications, such as surveillance, human-computer interaction, and healthcare monitoring.
By relying on complex spatio-temporal local encoding, most existing point cloud sequence networks capture spatio-temporal local structures to recognize 3D human actions.
To simplify the point cloud sequence modeling task, we propose a lightweight and effective point cloud sequence network referred to as SequentialPointNet for real-time 3D action recognition.
Instead of capturing spatio-temporal local structures, SequentialPointNet encodes the temporal evolution of static appearances to recognize human actions.
Firstly, we define a novel type of point data, Hyperpoint, to better describe the temporally changing human appearances.
A theoretical foundation is provided to clarify the information equivalence property for converting point cloud sequences into Hyperpoint sequences.
Secondly, the point cloud sequence modeling task is decomposed into a Hyperpoint embedding task and a Hyperpoint sequence modeling task.
Specifically, for Hyperpoint embedding, the static point cloud technology is employed to convert point cloud sequences into Hyperpoint sequences, which introduces inherent frame-level parallelism; for Hyperpoint sequence modeling, a Hyperpoint-Mixer module is designed as the basic building block to learning the spatio-temporal features of human actions.
Extensive experiments on three widely-used 3D action recognition datasets demonstrate that the proposed SequentialPointNet achieves competitive classification performance with up to 10X faster than existing approaches.
\end{abstract}


\section{Introduction}
\label{sec:introduction}
\IEEEPARstart{W}{ith} the release of low-cost depth cameras, 3D human action recognition has attracted more and more attention from researchers, which has broad industrial applications, such as human-computer interaction, security surveillance, automated driving, and robotics applications\cite{Y4}. 
3D human action recognition \cite{S6,D7,P1} can be divided into three categories based on different data types: 1) skeleton sequence-based 3D human action recognition, which focuses on human joint trajectories; 2) depth sequence-based 3D human action recognition, in which pixel values in depth maps represent the distance information from human bodies to the depth camera; 3) point cloud sequence-based 3D human action recognition, which studies the spatial distribution of human appearance along the time dimension.
Compared with skeleton sequence-based approaches, point cloud sequences are more convenient to be collected without additional pose estimation algorithms.
Compared with depth sequence-based approaches, point cloud sequence-based methods yield lower computation costs.
For these reasons, we focus on point cloud sequence-based 3D human action recognition in this work.
\begin{figure}[t]
	\centering
	\includegraphics[width=7.5cm]{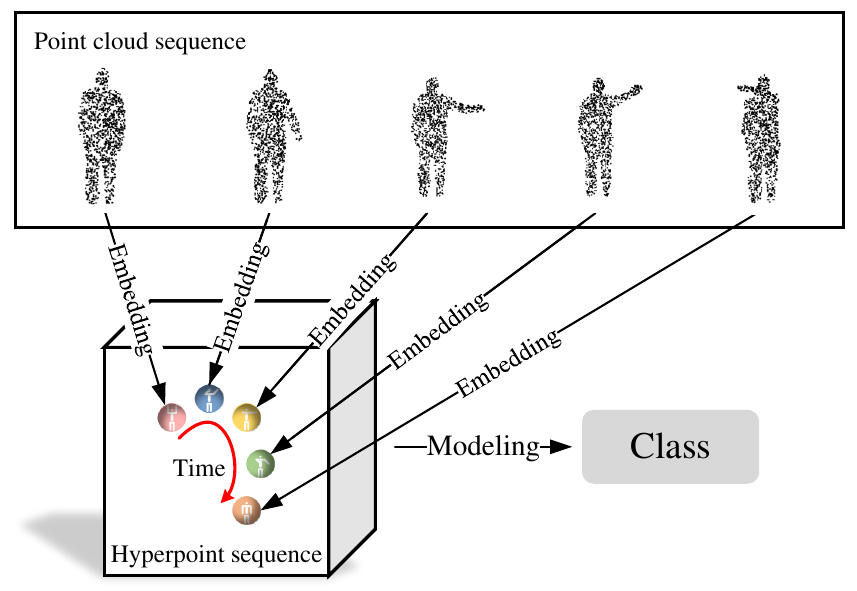}
	\caption{SequentialPointNet decomposes the complex point cloud sequence modeling task into a static point cloud technology-based Hyperpoint embedding task and a simple Hyperpoint sequence modeling task.}\label{fig1}
\end{figure}
\par Point cloud sequences of 3D human actions exhibit unordered intra-frame spatial information and ordered inter-frame temporal information.
Due to the complex data structures, capturing the spatio-temporal textures from point cloud sequences is extremely challenging. 
An intuitive way is to convert the point cloud sequence to a static point cloud and employ the static point cloud method to encode \cite{3DV}.
However, using static point clouds to represent the entire point cloud sequence loses a lot of spatio-temporal information, which degrades the recognition performance.
Therefore, point cloud sequence methods are necessary, which directly consume the point cloud sequence for 3D human action classification.
In order to model the dynamics of point cloud sequences, cross-frame spatio-temporal local neighborhoods are usually constructed.
After that, point spatio-temporal operations, such as 4D PointNet \cite{MeteorNet}, point spatio-temporal convolution \cite{PSTNet}, and point 4D convolution \cite{P4Transformer}, are used to encode the spatio-temporal local structures.
Nevertheless, unlike conventional grid-based videos, spatio-temporal local encoding for point cloud sequences is both sophisticated and time-consuming because point cloud sequences are irregular and unordered in the spatial dimension.
Furthermore, cross-frame spatio-temporal local encoding restricts frame-level parallel computation, which is not conducive to real-time point cloud sequence modeling.
\par In this paper, we propose a lightweight frame-level parallel point cloud sequence network, named SequentialPointNet, for real-time 3D human action recognition without resorting to spatio-temporal local encoding.
SequentialPointNet remarkably simplifies the complexity of the point cloud sequence modeling task and improves computational efficiency while achieving strong recognition performance.
SequentialPointNet encodes the temporal evolution of static appearances instead of capturing spatio-temporal local structures to recognize human actions.
We define a novel type of point data named Hyperpoint to better describe the temporally changing human appearances.
As shown in Fig. \ref{fig1}, the key to our approach is to decompose the complex point cloud sequence modeling task into a static point cloud technology-based Hyperpoint embedding task and a simple Hyperpoint sequence modeling task.
Specifically, we employ the static point cloud technology as a Hyperpoint embedding module to flatten each point cloud frame into a Hyperpoint.
In this fashion, the spatial structure of the human appearance is encoded and embedded into the corresponding Hyperpoint. 
Further, these Hyperpoints are assembled together and form a Hyperpoint sequence.
We design a Hyperpoint-Mixer module as the basic building block to model Hyperpoint sequences for recognizing 3D human actions.
\par Our main contributions are summarized as follows:
\par•	To avoid computationally expensive spatio-temporal local encoding, we propose an efficient and effective point cloud sequence network, dubbed SequentialPointNet, for real-time 3D human action recognition. 
SequentialPointNet treats point cloud sequence modeling as a two-phase task: Hyperpoint embedding and Hyperpoint sequence modeling, which significantly simplifies the encoding complexity and improves computational efficiency while realizing superior recognition performance.
\par•	To the best of our knowledge, we are the first to mathematically define Hyperpoint and design a Hyperpoint-Mixer module as the basic building block to model Hyperpoint sequences.
\par•	Our SequentialPointNet achieves up to 10$\times$ faster than existing point cloud sequence models and yields the cross-view accuracy of 97.6$\%$ on the NTU RGB+D 60 dataset, the cross-setup accuracy of 95.4$\%$ on the NTU RGB+D 120 dataset, and the cross-subject accuracy of 92.64$\%$ on the MSR Action3D dataset, which outperforms the state-of-the-art methods. Moreover, our SequentialPointNet adapted to skeleton sequences achieves better performance compared to the most famous skeleton sequence-based action recognition method.
\par This paper is an extension of our previous work HyperpointNet \cite{HyperpointNet}.
In HyperpointNet, to avoid spatio-temporal local encoding, the Hyperpoint is proposed as a new point data type but its mathematical definition is not provided.
HyperpointNet converts the point cloud sequence into a Hyperpoint sequence and further converts the Hyperpoint sequence into static point clouds.
By doing so, HyperpointNet has the ability to model point cloud sequences utilizing only static point cloud encoding techniques.
In SequentialPointNet, we provide a detailed mathematical definition and application scenarios of Hyperpoint sequence and compare it with existing point cloud data.
SequentialPointNet explicitly decomposes the point cloud sequence modeling task into a Hyperpoint embedding task and a Hyperpoint sequence modeling task, as well as gives a theoretical foundation for transforming point cloud sequences into Hyperpoint sequences.
For the Hyperpoint sequence modeling task, SequentialPointNet designs a specialized network architecture named Hyperpoint-Mixer for Hyperpoint sequences rather than transforming them into static point clouds. 
Experiments also demonstrate that SequentialPointNet is superior to HyperpointNet.
Compared to our previous work, we include another dataset for 3D action recognition, $i.e.$, MSR Action3D dataset\cite{MSR}.
Moreover, we further apply our SequentialPointNet to skeleton sequence-based 3D human action recognition on NTU RGB+D 60 dataset\cite{NTU60}, which verifies the effectiveness of SequentialPointNet's adaption to a skeleton sequence modeling task.
\par SequentialPointNet's code has been made public at \href{https://github.com/XingLi1012/SequentialPointNet.git}{https://github.com/XingLi1012/SequentialPointNet.git}.
\section{Related Work}
\subsection{Static Point Cloud Modeling}
With the popularity of low-cost 3D sensors, deep learning on static point clouds has attracted much attention from researchers due to extensive applications ranging from object classification, part segmentation, to scene semantic parsing.
Static point clouds models \cite{PVB5,PointNet} can be divided into volumetric-based methods and point-based methods. 
Volumetric-based methods usually voxelize a point cloud into 3D grids, and then a 3D Convolution Neural Network (CNN) is applied on the volumetric representation for classification.
Point-based methods are directly performed on raw point clouds.
PointNet \cite{PointNet} is a pioneering effort that directly processes point sets. 
The key idea of PointNet is to abstract each point using a set of Multi-layer Perceptrons (MLPs) and then assemble all individual point features by a symmetry function, $i.e$., a max pooling operation.
However, PointNet lacks the ability to capture local structures.
Therefore, in \cite{Pointnet++}, a hierarchical network PointNet++ is proposed to encode fine geometric structures from the neighborhood of each point.
PointNet++ is made of several set abstraction levels. 
A set abstraction level is composed of three layers: sampling layer, grouping layer, and PointNet-based learning layer. 
By stacking several set abstraction levels, PointNet++ progressively abstracts larger and larger local regions along the hierarchy.
\subsection{Point Cloud Sequence-based 3D Human Action Recognition}
Point cloud sequence-based 3D human action recognition is a fairly new and challenging task.
3DV-PointNet++ \cite{3DV} is the first volumetric-based work to recognize human actions from point cloud sequences.
In 3DV-PointNet++, 3D dynamic voxel (3DV) is proposed as a novel 3D motion representation.
A set of points is extracted from 3DV and input into PointNet++ for 3D action recognition in the end-to-end learning way.
However, since the point cloud sequence is converted into a static 3D point cloud set, 3DV-PointNet++ loses a lot of spatio-temporal information and increases additional computational costs.
\par To overcome this problem, researchers have focused mainly on investigating point cloud sequence networks that directly consume point cloud sequences for human action recognition.
MeteorNet \cite{MeteorNet} is the first work on deep learning for modeling point cloud sequences. 
In MeteorNet, two ways are proposed to construct spatio-temporal local neighborhoods for each point in the point cloud sequence. 
The abstracted features of each point are learned by aggregating the information from these neighborhoods.
Fan $et$ $al$. \cite{PSTNet} propose a Point spatio-temporal (PST) convolution to encode the spatio-temporal local structures of point cloud sequences.
PST convolution first disentangles space and time in point cloud sequences. 
Then, PST convolutions are incorporated into a deep network namely PSTNet to model point cloud sequences in a hierarchical manner. 
To avoid point tracking, Point 4D Transformer (P4Transformer) network \cite{P4Transformer} is proposed to model point cloud videos.
Specifically, P4Transformer consists of a point 4D convolution to embed the spatio-temporal local structures presented in a point cloud video and a Transformer to encode the appearance and motion information by performing self-attention on the embedded local features. 
However, in these point cloud sequence networks, spatio-temporal local encoding is usually performed during modeling point cloud sequences, which is quite time-consuming and limits the real-time computational efficiency.
In HyperpointNet\cite{HyperpointNet}, the point cloud sequence is first abstracted into a new point data type named Hyperpoint sequence.
Then, the Hyperpoint sequence is converted into static point clouds by injecting order information.
In this fashion, the point cloud sequence modeling task is decomposed into two static point cloud encoding tasks.
HyperpointNet is the first deep neural network to model point cloud sequences without performing cross-frame spatio-temporal local encoding.
However, HyperpointNet, only based on static point cloud technology, does not fully exploit the spatio-temporal information of the Hyperpoint sequence.
\begin{figure}[t]
	\centering
	\includegraphics[width=7cm]{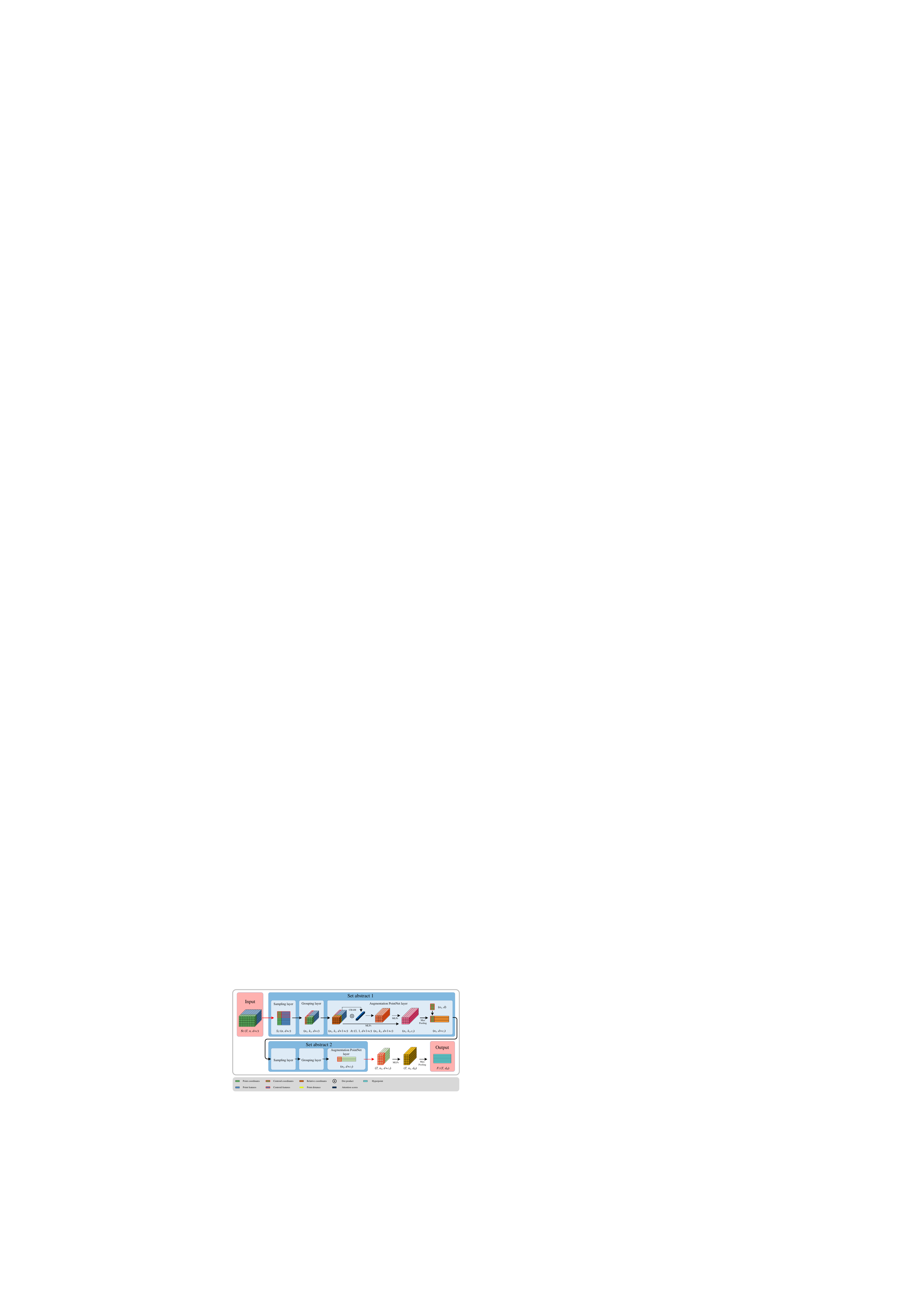}
	\caption{SequentialPointNet contains a Hyperpoint embedding module, a Hyperpoint-Mixer module, and a classifier head.}\label{fig2}
\end{figure}
\section{METHODOLOGY}
In this section, we present a lightweight and effective point cloud sequence network referred to as SequentialPointNet for real-time 3D action recognition.
Rather than capturing spatio-temporal local structures, our SequentialPointNet encodes the temporal evolution of static appearances to recognize human actions.  
By innovatively decomposing the modeling task of point cloud sequences into a Hyperpoint embedding task and a Hyperpoint sequence modeling task, SequentialPointNet significantly reduces the complexity and time cost for point cloud sequence modeling, while attaining very high recognition accuracy.
The overall flowchart of SequentialPointNet is described in Fig. \ref{fig2}.
SequentialPointNet contains a Hyperpoint embedding module, a Hyperpoint-Mixer module, and a classifier head.
\subsection{Hyperpoint Sequence}
We define a new type of point data named Hyperpoint.
A Hyperpoint sequence of length $T$ is represented as $\bm{F}=\left\{\bm{F}_{t} \mid t=1, \ldots, T\right\}$ with $\bm{F}_{t} \in \mathbb{R}^{d_H+m_H}$, which is an ordered set of high-dimensional points.
Each Hyperpoint $\bm{F}_{t}$ is a vector of its $d_H$-dim coordinate plus extra $m_H$-dim feature channels.
Extra feature channels can be neighborhood structure information or other modal features from different sensors.
In this paper, unless otherwise noted, we only use the high-dimensional coordinate as our Hyperpoint's channels.
\par The Hyperpoint sequence can be used to describe a temporally changing complex information unit, which exists widely in the real world such as human spatio-temporal actions, 4D driving scenarios, etc.
In this paper, the Hyperpoint sequence is obtained by flattening each frame of the point cloud sequence.
Each Hyperpoint encapsulates the complex spatial structure of the human appearance at a specific moment.
\par \textbf{Comparison with static point cloud and point cloud sequence:}
The static point cloud or individual point cloud frame is essentially an unordered point set.
Ordered point cloud frames further form a point cloud sequence.
Point cloud sequences are ordered in the temporal dimension but spatially irregular.
In contrast, Hyperpoint sequences do not include unordered point sets and are easier to be encoded.
Furthermore, unlike static point clouds and point cloud sequences, each element in the Hyperpoint sequence carries complex internal information.
The comparison of the three types of point data is presented in Fig. \ref{fig3}.  
\begin{figure}[t]
	\centering
	\includegraphics[width=7cm]{Fig3}
	\caption{The comparison of the three types of point data.}\label{fig3}
\end{figure}
\subsection{Hyperpoint Embedding Module}
To extract features of video data, spatio-temporal local encoding is usually performed. 
However, as point cloud sequences are irregular and lack order along the spatial dimension, spatio-temporal local encoding for point cloud sequences is complex and time-consuming.
Moreover, spatio-temporal local encoding limits real-time point cloud sequence classification due to heavy cross-frame computational dependencies.
\par To avoid spatio-temporal local encoding, SequentialPointNet disentangles space and time in point cloud sequences.
In our SequentialPointNet, the Hyperpoint embedding module is performed to preferentially encode each point cloud frame and output a Hyperpoint summarizing spatial structures.
After that, the complex point cloud sequence modeling task is transformed into a simple Hyperpoint sequence modeling task.
An outstanding advantage of this space-first strategy is that existing static point cloud models \cite{PointNet,Pointnet++} can be used almost out of the box as the Hyperpoint embedding module.
In addition, embedding each point cloud frame independently allows the Hyperpoint sequences to be generated in a frame-level parallel manner, thereby improving modeling efficiency.
\par In the Hyperpoint embedding module, each point cloud frame is transformed into a Hyperpoint summarizing human static appearance by the static point cloud technology.
We first adopt the set abstraction operation twice to downsample each point cloud frame.
In this process, the texture information from space partitions is aggregated into the corresponding centroids.
Then, in order to characterize human entire appearance information, a PointNet layer is used.
\par The set abstraction operation in our work is made of three key layers: sampling layer, grouping layer, and augmentation PointNet layer.
Specifically, let $\bm{{S}}=\left\{\mathcal{S}_{t}\right\}_{t=1}^{T}$ denote a point cloud sequence of $T$ frames, and $\mathcal{S}_{t}=\left\{x^t_1, x^t_2, ..., x^t_n  \right\}$ denotes the unordered point set of the $t$-th frame, where $n$ is the number of points.
The $m$-th set abstraction level takes an $n_{m-1} \times (d + c_{m-1})$ matrix as input that is from $n_{m-1}$ points with $d$-dim coordinates and $c_{m-1}$-dim point features. 
It outputs an $n_{m} \times\left(d+c_{m}\right)$ matrix of $n_{m}$ downsampled points with $d$-dim coordinates and new $c_{m}$-dim feature vectors summarizing local context.
The size of the input in the first set abstraction level is $n \times (d + c)$.
In this work, $d$ is set as 3 corresponding to the 3D coordinate ($X$, $Y$, $Z$) of each point, and $c$ is set as 0.
\par In the sampling layer, farthest point sampling (FPS)\cite{Pointnet++} is used to choose $n_{m}$ points as centroids from the point set.
\par In the grouping layer, a point set of size $n_{m-1} \times (d + c_{m-1})$ and the coordinates of a set of centroids of size $n_{m} \times d$ are taken as input. 
The output is $n_{m}$ groups of point sets of size $n_{m} \times k_{m} \times \left(d+c_{m-1}\right)$, where each group corresponds to a local region and $k_{m}$ is the number of local points in the neighborhood of centroid points.
Ball query finds all points that are within a radius to the query point, in which an upper limit of $k_{m}$ is set.
\par The augmentation PointNet layer in the set abstraction operation includes an inter-feature attention mechanism, a set of MLPs, and a max pooling operation.
The input of this layer is $n_{m}$ local regions with data size $n_{m} \times k_{m} \times \left(d+c_{m-1}\right)$.
First, the coordinates of points in a local region are translated into a local frame relative to the centroid point.
Second, the distance between each local point and the corresponding centroid is used as a 1D additional point feature to alleviate the influence of rotational motion on action recognition.
Then, an inter-feature attention mechanism is adopted to optimize the fusion effect of different features.
The inter-feature attention mechanism is realized by Channel Attention Module (CAM).
The inter-feature attention mechanism is not used in the first set abstraction operation due to only the 1-dim point feature.
In the following, a set of MLPs are applied to abstract the features of each local point.
Further, the representation of a local region is generated by incorporating the abstracted features of all local points using a max pooling operation.
Finally, coordinates of the centroid point and its local region representation are concatenated as abstracted features of this centroid point. 
The augmentation PointNet layer is formalized as follows:
\begin{equation}
{r}_{j}^{t}=\left[\underset{i=1, \ldots, k_{m}}{\operatorname{MAX}}\left\{\operatorname{MLP}\left(\left[\left({l}_{j,i}^{t}-{o}_{j}^{t}\right);{e}_{j,i}^{t};{p}_{j,i}^{t}\right]\odot A\right)\right\};{o}_{j}^{t}\right]
\end{equation}
where ${l}_{j,i}^{t}$ is the coordinates of $i$-th point in the $j$-th local region from the $t$-th point cloud frame. ${o}_{j}^{t}$ and ${p}_{j,i}^{t}$ are the coordinates of the centroid point and the point features corresponding to ${l}_{j,i}^{t}$, respectively. 
${e}_{j,i}^{t}$ is the euclidean distance between ${l}_{j,i}^{t}$ and ${o}_{j}^{t}$.
$A$ denotes the attention mechanism with (3+1+$c_{m-1}$)-dim scores corresponding to the coordinates and features of each point.
Attention scores in $A$ are shared among all local points from all point cloud frames.
$\odot$ indicates dot product operation.
${r}_{j}^{t}$ is the abstracted features of the $j$-th centroid point from the $t$-th point cloud frame. 
\par The set abstract operation is performed twice in the Hyperpoint sequence module.
In order to characterize the spatial appearance information of the entire point cloud frame, a PointNet layer consisting of a set of MLPs and a max pooling operation is used as follows:
\begin{equation}
\bm{F}_{t}=\underset{j=1, \ldots, n_2}{\operatorname{MAX}}\left\{\operatorname{MLP}\left({r}_{j}^{t}\right)\right\}
\end{equation}
where $\bm{F}_{t}$ is the Hyperpoint of the $t$-th point cloud frame.
So the Hyperpoint sequence is represented as $\bm{{F}}=\left\{\bm{F}_{t}\right\}_{t=1}^{T}$.
\par \textbf{Theoretical foundation for converting point cloud sequences to Hyperpoint sequences:}
Temporally changing static appearances constitute the human spatio-temporal action.
Our SequentialPointNet encodes the temporal evolution of static appearances instead of capturing spatio-temporal local structures to distinguish human actions.
In our work, point cloud sequences are converted into Hyperpoint sequences to record the static appearances at each moment.
To demonstrate the information equivalence property between point cloud sequences and Hyperpoint sequences, we provide a theoretical foundation for our converting operation by showing the universal approximation ability of the Hyperpoint embedding module to continuous functions on point cloud frames.
\begin{figure*}[t]
	\centering
	\includegraphics[width=15cm]{Fig4}
	\caption{Left: Hyperpoint-Mixer module. Right: Space dislocation layer.}\label{fig4}
\end{figure*}
\par Formally, let $\mathcal{X}=\left\{S: S \subseteq[0,1]^{c}\right.$ and $\left.|S|=n\right\}$ is the set of c-dimensional point clouds inside a c-dimensional unit cube. $f:\mathcal{X} \rightarrow \mathbb{R}$ is a continuous set function on $\mathcal{X}$ w.r.t to Hausdorff distance $D_{H}(\cdot, \cdot)$, i.e., $\forall \epsilon>0, \exists \delta>0$, for any $S, S^{\prime} \in \mathcal{X}$, if $D_{H}(\cdot, \cdot)<\delta$, then $|f(S)-f(S^{\prime})|<\epsilon$. 
The theorem 1 \cite{PointNet} says that $f$ can be arbitrarily approximated by PointNet given enough neurons at the max pooling layer.
\par $\textbf{Theorem 1.}$ $\textit{Suppose}$ $f: \mathcal{X} \rightarrow \mathbb{R}$ $\textit{is}$ $\textit{a}$ $\textit{continuous}$ $\textit{set}$ $\textit{function}$ $\textit{w.r.t}$ $\textit{Hausdorff}$ $\textit{distance}$ $D_{H}(\cdot, \cdot) . \quad \forall \epsilon>$ $0, \exists$ $\textit{a}$ $\textit{continuous}$ $\textit{function}$ $h$ $\textit{and}$ $\textit{a}$ $\textit{symmetric}$ $\textit{function}$ $g\left(x_{1}, \ldots, x_{n}\right)=\gamma \circ M A X$$\textit{, such}$ $\textit{that}$ $\textit{for}$ $\textit{any}$ $S \in \mathcal{X}$,
$$
\left|f(S)-\gamma\left(\underset{x_{i} \in S}{\operatorname{MAX}}\left\{h\left(x_{i}\right)\right\}\right)\right|<\epsilon
$$
$\textit{where}$ $x_{1}, \ldots, x_{n}$ $\textit{is}$ $\textit{the}$ $\textit{full}$ $\textit{list}$ $\textit{of}$ $\textit{elements}$ $\textit{in}$ $S$ $\textit{ordered}$ $\textit{arbitrarily,}$ $\gamma$ $\textit{is}$ $\textit{a}$ $\textit{continuous}$ $\textit{function,}$ $\textit{and}$ $\textit{MAX}$ $\textit{is}$ $\textit{a}$ $\textit{vector}$ $\textit{max}$ $\textit{operator}$ $\textit{that}$ $\textit{takes}$ $n$ $\textit{vectors}$ $\textit{as}$ $\textit{input}$ $\textit{and}$ $\textit{returns}$ $\textit{a}$ $\textit{new}$ $\textit{vector}$ $\textit{of}$ $\textit{the}$ $\textit{element-wise}$ $\textit{maximum}$.
\par As stated above, continuous functions can be arbitrarily approximated by PointNet given enough neurons at the max pooling layer.
The Hyperpoint embedding module is a recursive application of PointNet on nested partitions of the input point set.
Therefore, the Hyperpoint embedding module is able to arbitrarily approximate continuous functions on point cloud frames given enough neurons at max pooling layers and a suitable partitioning strategy.
In other words, the Hyperpoint embedding module is capable of extracting the complete static appearance information from point cloud frames to generate Hyperpoint sequences of equivalent information.
\subsection{Hyperpoint-Mixer Module}
Hyperpoint-Mixer module is proposed to model Hyperpoint sequence for 3D action recognition.
The Hyperpoint sequence is an ordered set of high-dimensional single points.
Compared to static point clouds and point cloud sequences, modeling Hyperpoint sequences is extremely simple due to its ordered elements.
In the Hyperpoint-Mixer module, we first dislocate Hyperpoints to specific temporal marking regions.   
Then, the dimension-mixing operation and the Hyperpoint-mixing operation are separately conducted to generate the final human spatio-temporal features.
By doing so, the spatial and temporal information are decoupled, minimizing the impact of spatial irregularity on temporal information.
Fig. \ref{fig4} summarizes the Hyperpoint-Mixer module.
The proposed Hyperpoint-Mixer module consists of multiple space dislocation layers of identical size, a multi-level feature learning based on skip-connection operations, and a Hyperpoint-mixing ($\mathcal{H}$-MIX) layer.

\subsubsection{Space dislocation layer}
The space dislocation layer is composed of a Hyperpoint dislocation block and a dimension-mixing ($\mathcal{D}$-MIX) block, which is presented to learn the internal spatial feature of dislocated Hyperpoints. 
\par In the Hyperpoint dislocation block, Hyperpoints are spatially dislocated to record the temporal order by adding the corresponding temporal order vectors.
The essence of the Hyperpoint dislocation block is to dislocate each Hyperpoint to a unique temporal marking region.
In this fashion, the new coordinate of the dislocated Hyperpoint can be considered as the sum of two vectors: the temporal order vector and the spatial structure vector. The temporal order vector is used to distinguish the temporal order of the Hyperpoints while the spatial structure vector represents the internal appearance structure.
Temporal order vectors ($ToV$) can be generated using the sine and cosine functions of different frequencies \cite{Transformer}:
\begin{equation}
ToV_{t, 2 h}=\sin \left(t/ 10000^{2 h / d_{\text {H }}}\right)
\end{equation}
\begin{equation}
ToV_{t, 2 h+1}=\cos \left(t/ 10000^{2 h / d_{\text {H}}}\right)
\end{equation}
where $d_{\text {H}}$ denotes the dimension number of Hyperpoint coordinates. $t$ is the temporal position and $h$ is the dimension position.
\par Then, $\mathcal{D}$-MIX block learns the internal appearance structure of dislocated Hyperpoints in different temporal marking region, maps $\mathbb{R}^{d_\text{H}}\rightarrow \mathbb{R}^{d_\text{H}}$, and is shared across all dislocated Hyperpoints. Each $\mathcal{D}$-MIX block contains a set of MLP operations, a batch norm, and a ReLU non-linearity. 
Space dislocation layers can be formalized as follows:
\begin{equation}
\bm{F}^{\ell}_{t}=\mathcal{D}\operatorname{-MIX}(\bm{F}^{\ell-1}_{t}+ToV_t), \quad \text { for } t=1 \ldots T.
\end{equation}
where $\bm{F}^{\ell}$ is the new Hyperpoint sequences after the $\ell$-th space dislocation layer.
\par Each space dislocation layer takes an input of the same size, which is an isotropic design. 
With the stacking of space dislocation layers, Hyperpoints are dislocated at increasingly larger scales.
In the larger dislocation space, the coordinates of Hyperpoints record more temporal information but less spatial information. 	
To facilitate network optimization, multi-level features from different dislocation spaces are added by skip-connection operations and sent into $\mathcal{H}$-MIX layer as follows:
\begin{equation}
\bm{R}_{i}=\underset{t=1, \ldots, T}{\mathcal{H}\operatorname{-MIX}}\left\{(\bm{F}+\bm{F}^1+,...,+\bm{F}^{\ell})_{t,i}\right\}
\end{equation}
where $\bm{R}$ is the spatio-temporal feature of the Hyperpoint sequence.
\subsubsection{Hyperpoint-mixing layer}
$\mathcal{H}$-MIX layer is applied to aggregate spatial structures from all the Hyperpoints for the final spatio-temporal features. 
Since the temporal information has been injected, we can only use the simple max pooling to gather spatial information from the temporal marking regions.
Benefiting from the order property of Hyperpoint sequences, temporal partitions can be easily divided.
In order to capture the subactions within the Hyperpoint sequence, the hierarchical pyramid max pooling operation is adopted as the $\mathcal{H}$-MIX layer, which divides the fused Hyperpoint sequence $\bm{\widetilde{F}}=\bm{F}+\bm{F}^1+,...,+\bm{F}^{\ell}$ into multiple temporal partitions of the equal number of Hyperpoints and then performs the max pooling operation in each partition to generate corresponding spatio-temporal features.
In this work, we employ a 2-layer pyramid with three partitions.
Spatio-temporal features from all temporal partitions are simply concatenated to form the final spatio-temporal feature $\bm{R}$.
Finally, the output of the Hyperpoint-Mixer module is input to a fully-connected classifier head for recognizing human actions.
\section{Experiments}
In this section, we firstly introduce the datasets. 
Then, we compare our SequentialPointNet with the existing state-of-the-art methods.
Again, we conduct detailed ablation studies to further demonstrate the contributions of different components in our SequentialPointNet.
After that, we compare the memory usage and computational efficiency of our SequentialPointNet with other point cloud sequence models.
Finally, we apply our SequentialPointNet to skeleton sequence-based 3D human action recognition on NTU RGB+D 60 dataset.

\subsection{Datasets}
\par We evaluate the proposed method on two large-scale public datasets ($i.e.$, NTU RGB+D 60 \cite{NTU60} and NTU RGB+D 120 \cite{NTU120}) and a small-scale public dataset ($i.e.$, MSR Action3D dataset \cite{MSR}).
\par The NTU RGB+D 60 dataset is composed of 56880 depth video sequences and skeleton video sequences for 60 actions and is one of the largest human action datasets. Both cross-subject and cross-view evaluation criteria are adopted for training and testing.
\par The NTU RGB+D 120 dataset is the largest dataset for 3D action recognition, which is an extension of the NTU RGB-D 60 dataset. The NTU RGB+D 120 dataset is composed of 114480 depth video sequences for 120 actions. Both cross-subject and cross-setup evaluation criteria are adopted for training and testing.
\par The MSR Action3D dataset contains 557 depth video samples of 20 actions from 10 subjects. Each action is performed 2 or 3 times by every subject. We adopt the same cross-subject settings in \cite{MSR}, where all the 20 actions are employed. Half of the subjects are used for training and the rest for testing.
\begin{table}[t]
	
	\renewcommand{\arraystretch}{1.0}
	\caption{Action recognition accuracy ($\%$) on NTU RGB+D 60}
	\label{table1}
	\centering
	\begin{tabular*}{\hsize}{@{}@{\extracolsep{\fill}}cccc@{}}
		
		\hline
		\bfseries Method$/$Year & \bfseries Input & \bfseries Cross-subject& \bfseries Cross-view\\
		
		\hline
		\noalign{\global\arrayrulewidth1pt}
		\noalign{\global\arrayrulewidth0.4pt}
		Wang $et$ $al$.(2018)\cite{DPBL} & depth & 87.1& 84.2\\
		MVDI(2019)\cite{MVDI} & depth & 84.6& 87.3\\
		3DFCNN(2020)\cite{3DFCNN} & depth & 78.1& 80.4\\
		Stateful ConvLSTM(2020)\cite{Statefull} & depth & 80.4& 79.9\\
		ST-GCN(2018)\cite{STGCN} & skeleton & 81.5& 88.3\\
		AS-GCN(2019)\cite{ASGCN} & skeleton & 86.8& 94.2\\
		SGN(2020)\cite{SGN} & skeleton & 89& 94.5\\
		3s-CrosSCLR(2021)\cite{3sCrosSCLR} & skeleton & 86.2& 92.5\\
		Sym-GNN(2021)\cite{symGNN} & skeleton & 90.1& 96.4\\
		3DV-PointNet++(2020)\cite{3DV}  & point& 88.8& 96.3\\
		P4Transformer(2021)\cite{P4Transformer} & point & 90.2& 96.4\\
		PSTNet(2021)\cite{PSTNet} & point  & $\mathbf{90.5}$& 96.5\\
		HyperpointNet(2022)\cite{HyperpointNet} & point  &90.2& 97.3\\
		\hline
		SequentialPointNet(ours) & point & 90.3& $\mathbf{97.6}$\\
		
		\hline
	\end{tabular*}
\end{table}
\begin{table}[t]
	\renewcommand{\arraystretch}{1.0}
	\caption{Action recognition accuracy ($\%$) on NTU RGB+D 120}
	\label{table2}
	\centering
	\begin{tabular*}{\hsize}{@{}@{\extracolsep{\fill}}cccc@{}}
		
		\hline
		\bfseries Method$/$Year & \bfseries Input & \bfseries Cross-subject& \bfseries Cross-setup\\
		
		\hline
		\noalign{\global\arrayrulewidth1pt}
		\noalign{\global\arrayrulewidth0.4pt}
		Baseline(2018)\cite{NTU120} & depth & 48.7& 40.1\\
		ST-GCN(2018)\cite{STGCN} & skeleton & 81.5& 88.3\\
		MS-G3D Net (2020)\cite{MSG3D} & skeleton & 86.9& 88.4\\
		4s Shift-GCN(2020)\cite{4sShiftGCN} & skeleton & 85.9& 87.6\\
		SGN(2020)\cite{SGN} & skeleton & 79.2& 81.5\\
		3s-CrosSCLR(2021)\cite{3sCrosSCLR} & skeleton & 80.5& 80.4\\
		3DV-PointNet++(2020)\cite{3DV}  & point& 82.4& 93.5\\
		P4Transformer(2021)\cite{P4Transformer} & point & 86.4& 93.5\\
		PSTNet(2021)\cite{PSTNet} & point  & $\mathbf{87.0}$& 93.5\\
		HyperpointNet(2022)\cite{HyperpointNet} & point  &83.2& 95.1\\
		\hline
		SequentialPointNet(ours) & point & 83.5& $\mathbf{95.4}$\\
		
		\hline
	\end{tabular*}
\end{table}
\subsection{Comparison with the State-of-the-art Methods}
In this section, in order to verify the recognition accuracy of our SequentialPointNet, comparison experiments with other state-of-the-art approaches are implemented on NTU RGB+D 60 dataset, NTU RGB+D 120 dataset,  and MSR Action3D dataset.
\subsubsection{NTU RGB+D 60 dataset}
We first compare our SequentialPointNet with the state-of-the-art methods on the NTU RGB+D 60 dataset.
As indicated in Table \ref{table1}, SequentialPointNet has recognition accuracy of 90.3$\%$ and 97.6$\%$ on the cross-subject and cross-view test settings, respectively.
SequentialPointNet shows strong performance on par or even better than other point sequence-based approaches.
Our SequentialPointNet achieves state-of-the-art performance among all methods on the cross-view test setting and results in similar recognition accuracy as PSTNet on the cross-subject test setting.
The key success of our SequentialPointNet lies in the effective encoding for the temporal appearance evolution by Hyperpoint-Mixer module and minimizing the impact of the spatial irregularity on temporal information.
Compared with our previous work HyperpointNet\cite{HyperpointNet} that transforms Hyperpoint sequences into static point clouds, Hyperpoint-Mixer module in our SequentialPointNet designs isotropic space dislocation layers to blend the spatial and temporal information under multiple scales, enhancing the feature extraction ability for Hyperpoint sequences.

\subsubsection{NTU RGB+D 120 dataset}
We then compare our SequentialPointNet with the state-of-the-art methods on the NTU RGB+D 120 dataset.
As indicated in Table \ref{table2}, SequentialPointNet achieves accuracy of 83.5$\%$ and 95.4$\%$ on the cross-subject and cross-setup test settings, respectively.
Compared with PSTNet, SequentialPointNet does not show a competitive recognition accuracy on the cross-subject setting.
However, SequentialPointNet gains a strong lead on the cross-setup test setting and achieves the highest recognition accuracy.
Moreover, compared to existing point cloud sequence models, our approach greatly simplifies the point cloud sequence modeling task by decomposing it into a Hyperpoint embedding task and a Hyperpoint sequence encoding task.
\begin{table}[t]
	\renewcommand{\arraystretch}{1.0}
	\caption{Action recognition accuracy ($\%$) on MSR-Action3D}
	\label{table3}
	\centering
	\begin{tabular*}{\hsize}{@{}@{\extracolsep{\fill}}cccc@{}}
		
		\hline
		\bfseries Method$/$Year & \bfseries Input & \bfseries  $\#$ Frames& \bfseries Accuracy\\
		
		\hline
		\noalign{\global\arrayrulewidth1pt}
		\noalign{\global\arrayrulewidth0.4pt}
		Kläser $et$ $al$.(2008)\cite{ASTDB}  & depth& 18& 81.43\\
		Vieira $et$ $al$.(2012)\cite{SSTO}  & depth& 20& 78.20\\
		MeteorNet(2019)\cite{MeteorNet}  & point& 24& 61.61\\
		PointNet++(2020)\cite{Pointnet++}  & point& 1& 88.50\\
		P4Transformer(2021)\cite{P4Transformer}  & point& 24& 90.94\\
		PSTNet(2021)\cite{PSTNet}  & point& 24& 91.20\\
		HyperpointNet(2022)\cite{HyperpointNet}  & point& 24& 91.54\\
		\hline
		SequentialPointNet(ours)  & point& 24&  $\mathbf{92.64}$\\		
		\hline
	\end{tabular*}
\end{table}
\subsubsection{MSR Action3D dataset}
In order to comprehensively evaluate our method, comparative experiments are also carried out on the small-scale MSR Action3D dataset.
To alleviate the overfitting problem on the small-scale dataset, the batch size is set as 16.
Other parameter settings remain the same as those on the two large-scale datasets.
Table \ref{table3} illustrates the recognition accuracy of different methods.
When using 24 point cloud frames as inputs, our model yields state-of-the-art performance on the MSR Action3D dataset.
Experimental results on the small-scale dataset demonstrate that our approach can achieve superior recognition accuracy even without a large amount of data for training.
\subsection{Ablation Study}
In this section, comprehensive ablation studies are performed on NTU RGB+D 60 dataset to validate the contributions of different components in our SequentialPointNet.
\subsubsection{Effectiveness of Hyperpoint-Mixer module}
We conduct the experiments to demonstrate the effectiveness of the Hyperpoint-Mixer module, and results are reported in Table \ref{table4}.
Several strong deep networks are used to instead of the Hyperpoint-Mixer module in our SequentialPointNet.
In SequentialPointNet (LSTM), LSTM is employed.
In SequentialPointNet (GRU), GRU is employed.
In SequentialPointNet (Transformer), a Transformer of two attention layers is used.
In SequentialPointNet (MLP-Mixer), a MLP-Mixer of two mixer layers is used.
\begin{table}[t]
	\renewcommand{\arraystretch}{1.0}
	\caption{Cross-view recognition accuracy ($\%$) when using models}
	\label{table4}
	\centering
	\begin{tabular*}{\hsize}{@{}@{\extracolsep{\fill}}cc@{}}
		
		\hline
		\bfseries Method & \bfseries Accuracy  \\
		
		\hline
		SequentialPointNet(LSTM) & 85.9\\
		SequentialPointNet(GRU) & 86.4\\
		SequentialPointNet(Transformer) & 81.6\\
		SequentialPointNet(MLP-Mixer) & 94.5\\
		\hline
		SequentialPointNet & 97.6\\

		\hline
	\end{tabular*}
\end{table}
\par We can see from Table \ref{table4} that results of SequentialPointNet (LSTM) and SequentialPointNet (GRU) are much worse when compared with SequentialPointNet.
Hyperpoint sequences can be regarded as a kind of time series data.
Different from the conventional time series data, the internal structures of elements rather than the element changes generate the main discriminant information.
Thus, time series models that perform strict time-varying reasoning are not applicable to Hyperpoint sequences.
Recently, self-attention-based Transformer has remained dominant in natural language processing\cite{Transformer} and computer vision \cite{ViT}.
However, due to the lack of larger-scale data for pre-training, SequentialPointNet (Transformer) does not show a competitive result.
SequentialPointNet (MLP-Mixer) achieves the accuracy of 94.5$\%$, which is 3.1$\%$ lower than our SequentialPointNet.
The reason for this is that channel-mixing and token-mixing are performed alternatively, resulting in the mutual influence between spatial and temporal information.
\subsubsection{Different temporal information embedding manners}
The Hyperpoint dislocation block in the Hyperpoint-Mixer module is employed to embed the temporal information. 
To demonstrate its effectiveness, we also report the results of SequentialPointNet (4D, w/o hdb), which does not use the Hyperpoint dislocation block and injects the order information by appending the 1D temporal dimension to raw 3D points in each point cloud frame. 
Results are tabulated in Table \ref{table5}.
From the table, we observe that SequentialPointNet with the Hyperpoint dislocation block outperforms SequentialPointNet (4D, w/o hdb).
Therefore, the temporal information embedding manner used in SequentialPointNet is more efficient.
From the experimental results, we can draw a conclusion that premature embedding of temporal information will affect spatial information encoding and decrease the recognition accuracy.
\begin{table}[t]
	\renewcommand{\arraystretch}{1.0}
	\caption{Cross-view recognition accuracy ($\%$) of different methods}
	\label{table5}
	\centering
	\begin{tabular*}{\hsize}{@{}@{\extracolsep{\fill}}cc@{}}
		
		\hline
		\bfseries Method & \bfseries Accuracy  \\
		
		\hline
		SequentialPointNet(4D, w/o hdb) & 95.4\\
		SequentialPointNet(w/o mlfl) & 96.9\\
		\hline
		SequentialPointNet & 97.6\\
		
		\hline
	\end{tabular*}
\end{table}
\subsubsection{Effectiveness of multi-level feature learning in the Hyperpoint-Mixer module}
With the stacking of the space dislocation layer in the Hyperpoint-Mixer module, Hyperpoints are dislocated at increasingly larger scales.
In the larger dislocation space, the coordinates of Hyperpoints record more temporal information but less spatial information. 	
In order to obtain more discriminant information, multi-level features from different dislocation spaces are added by skip-connection operations.
To verify the effectiveness of the multi-level feature learning, we report the result of SequentialPointNet (w/o mlfl) that classify human actions without the multi-level features.
We can see from Table \ref{table5} that the recognition accuracy of SequentialPointNet (w/o mlfl) decreases by 0.7$\%$ without the multi-level feature learning.
\subsection{Modeling Efficiency Analysis}
In this section, We first show the frame-level parallelism of SequentialPointNet by its computation graph.
Then, we evaluate the memory usage and computational efficiency of our method.
\subsubsection{Computation graph of SequentialPointNet}
\par
\begin{figure}[t]
	\centering
	\includegraphics[width=7cm]{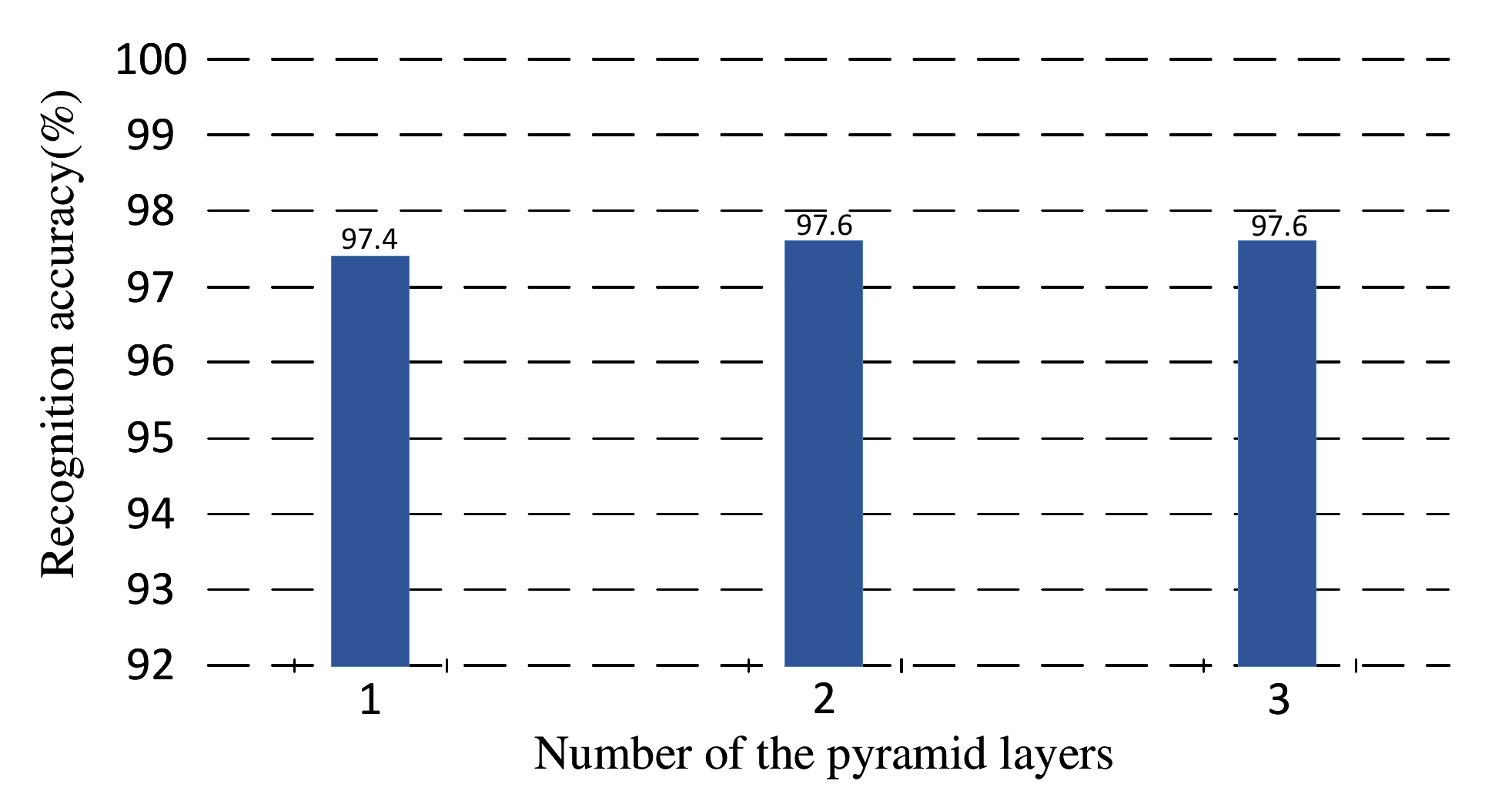}
	\caption{The computation graph of SequentialPointNet.\label{fig5}}
\end{figure}

 Fig. \ref{fig5} demonstrates the computation graph of SequentialPointNet.
Taking the Hyperpoint-mixing layer as the watershed, SequentialPointNet can be divided into a front part and a back part.
Since there is no cross-frame computational dependency, operations of the front part can be divided into frame-level units executed in parallel.
Each frame-level unit includes a Hyperpoint embedding module and all space dislocation layers. 
The back part only contains the network architectures of low computational complexity including the Hyperpoint-mixing operation and a classifier head.
Therefore, the main operations ($i.e.$, the front part) in SequentialPointNet can be executed in a frame-level parallel manner, based on which the modeling efficiency is greatly improved.
\subsubsection{Memory usage and computational efficiency}
Experiments are conducted on the machine with one Intel(R) Xeon(R) W-3175X CPU and one Nvidia RTX 3090 GPU.
In Table \ref{table6}, the number of parameters and the running time of our SequentialPointNet are compared with MeteorNet, 3DV-PointNet++, P4Transformer, and PSTNet.
The running time is the network forward inference time per point cloud sequence.

\par From the table, we can see that parameters in 3DV-PointNet++ are the fewest in all methods.
This is because 3DV-PointNet++ converts point cloud sequences to 3D point clouds, and employs static point cloud methods to process 3D point clouds.
Compared with point cloud sequence methods, the static point cloud method has fewer parameters but lower recognition accuracy.
The parameter number of SequentialPointNet is much less than MeteorNet, P4Transformer, and PSTNet, which verifies that our approach is the most lightweight point cloud sequence model.
In addition, SequentialPointNet is far faster than other methods.
SequentialPointNet takes only 1.32 milliseconds to classify a point cloud sequence far beyond real-time requirement.
In 3DV-PointNet++, the multi-stream network limits its parallelism.
In MeteorNet, P4Transformer, and PSTNet, spatio-temporal local encoding are performed, which is time-consuming and not conducive to parallel computing.   
SequentialPointNet improves the speed of point cloud sequence modeling by more than 10 times.
The superior computational efficiency of our SequentialPointNet is due to the lightweight network architecture and strong frame-level parallelism.
\par Additionally, our SequentialPointNet facilitates computational flexibility for point cloud sequence modeling.
Since there is no computational dependency, in the case of limited computing power, frame-level units can also be deployed on different devices or executed sequentially on a single device.
\begin{table}[t]
	\renewcommand{\arraystretch}{1.0}
	\caption{Parameters and running times comparison}
	\label{table6}
	\centering
	\begin{tabular*}{\hsize}{@{}@{\extracolsep{\fill}}ccc@{}}
		
		\hline
		\bfseries Method  & \bfseries Parameters (M) &  \bfseries Time (ms)\\
		
		\hline
		\noalign{\global\arrayrulewidth1pt}
		\noalign{\global\arrayrulewidth0.4pt}
		MeteorNet\cite{MeteorNet} & 17.60&  33.56\\
		3DV-PointNet++\cite{3DV} & $\mathbf{1.24}$ &54.85\\
		P4Transformer\cite{P4Transformer} & 44.10& 17.58\\
		PSTNet\cite{PSTNet} & 8.44& 27.56\\
		\hline
		SequentialPointNet & 3.72& $\mathbf{1.32}$\\
		
		\hline
	\end{tabular*}
\end{table}
\begin{table}[t]
	\renewcommand{\arraystretch}{1.3}
	\caption{Comparison between SequentialPointNet and ST-GCN on NTU RGB+D 60}
	\label{table7}
	\centering
	\begin{tabular*}{\hsize}{@{}@{\extracolsep{\fill}}cccc@{}}
		
		\hline
		\bfseries Method  & \bfseries Parameters (M) &  \bfseries Time (ms) &  \bfseries Accuracy ($\%$)\\
		
		\hline
		\noalign{\global\arrayrulewidth1pt}
		\noalign{\global\arrayrulewidth0.4pt}
		ST-GCN\cite{STGCN} & 3.12& 4.62$\times 10^-1$& 86.63\\
		SequentialPointNet & 0.55& 1.84$\times 10^-2$& 88.96\\
		
		\hline
	\end{tabular*}
\end{table}
\subsection{SequentialPointNet for Skeleton Sequence-based 3D Human Action Recognition}
SequentialPointNet can remarkably simplify the complexity of the sequence video data modeling task by decomposing it into a Hyperpoint embedding task based on corresponding existing static data encoding technology and a generic Hyperpoint sequence encoding task.
To verify the effectiveness of SequentialPointNet's adaption to a skeleton sequence modeling task, we further apply our method to skeleton sequence-based 3D human action recognition on NTU RGB+D 60 dataset\cite{NTU60}.
\par The frame in the skeleton sequence is a kind of graph structure data, representing the static appearance of the human body.
Therefore, for the skeleton sequence, the existing graph convolution network (GCN) layers\cite{STGCN} is used to construct a spatial graph convolution network as the Hyperpoint embedding module of SequentialPointNet.
The Hyperpoint-Mixer module remains the same as on the point cloud sequence.
For a fair comparison, we sample 24 frames with equal intervals from a skeleton video sequence. 
We compare our SequentialPointNet with the spatial-temporal graph convolutional network (ST-GCN)\cite{STGCN}, which is the most famous GCN-based method for skeleton sequence-based action recognition.
In Table \ref{table7}, the number of parameters, the running time, and the cross-view accuracy of our SequentialPointNet are compared with ST-GCN.
Skeleton sequences have a joint tracking nature that is convenient for exploring the joint-level movement information.
Note that, SequentialPointNet without utilizing the joint tracking nature achieves even better recognition performance than ST-GCN.
This is because that SequentialPointNet effectively encodes the temporally changing human appearances and minimizes the impact of spatial irregularity on temporal information.
Furthermore, SequentialPointNet yields fewer network parameters and faster running time than ST-GCN. 
\section{Conclusion}
In this paper, we propose a novel network named SequentialPointNet to model point cloud sequences for real-time 3D human action recognition. 
Instead of capturing spatio-temporal local structures, SequentialPointNet models the temporal evolution of static appearances to recognize human actions.
In our SequentialPointNet, the point cloud sequence modeling is treated as a two-phase task: Hyperpoint embedding and Hyperpoint sequence modeling.
We propose Hyperpoint as a new type of point cloud data and design a Hyperpoint-Mixer module as the basic building block to model Hyperpoint sequences for recognizing 3D human actions.
Extensive experiments conducted on three public datasets show that SequentialPointNet obtains superior recognition performance and improves the speed of point cloud sequence classification by more than 10 times, significantly facilitating real-time 3D human action recognition based on point cloud sequences.

\begin{IEEEbiography}[{\includegraphics[width=1in,height=1.25in,clip,keepaspectratio]{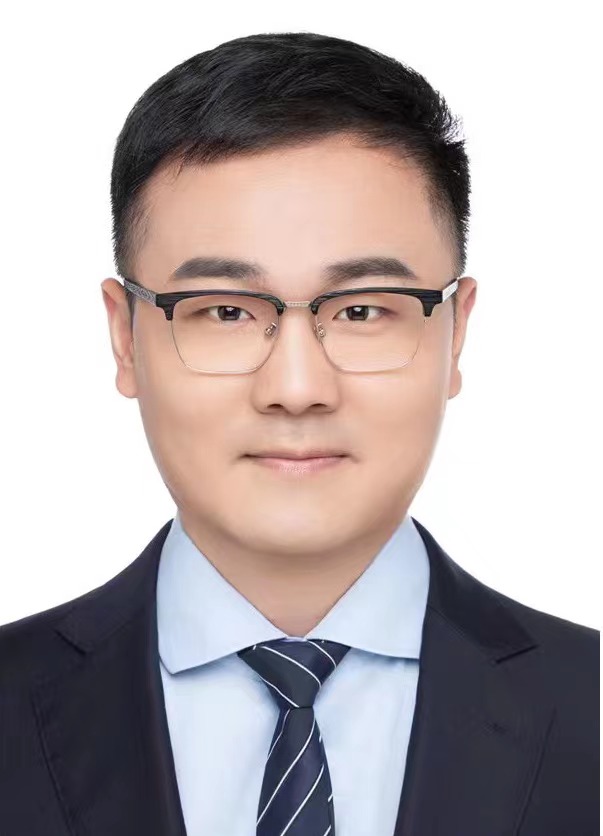}}]{Xing Li}
	received the B.Sc. and M.Sc. degrees in software engineering from Changzhou University, China, in 2016 and 2019, respectively. He is currently pursuing the Ph.D. degree with the Department of Computer Science and Technology, Hohai University, Nanjing, China. His current research interests include machine learning, computer vision and deep learning, especially human action recognition.
\end{IEEEbiography}
\begin{IEEEbiography}[{\includegraphics[width=1in,height=1.25in,clip,keepaspectratio]{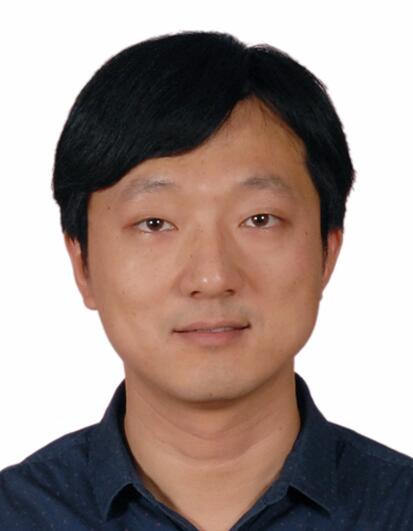}}]{Qian Huang}
	received the B. Sc. degree in computer science from Nanjing University, China, in 2003, and the Ph. D. degree in computer science from the Institute of Computing Technology, Chinese Academy of Sciences, in 2010. From 2010 to 2012, he was a deputy technical manager of Mediatek (Beijing) Incorporation, Beijing, China. Since Dec. 2012, he is with Hohai University, Nanjing, China, where he serves as the dean of Computer Science $\&$ Technology Department. His research interests lie in industry-specific multimedia computing, especially on video compression $\&$ communication, object identification $\&$ tracking, and behavior expression $\&$ analysis. He is a member of the CAAI Technical Committee on Deep Learning, a member of the CCF Technical Committee on Multimedia Technology, and a member of the CSIG Technical Committee on Multimedia. Currently he serves as an associate editor for IET Image Processing, and a reviewer for some IEEE Transactions such as TIP, TMM and TCSVT. 
\end{IEEEbiography}
\begin{IEEEbiography}[{\includegraphics[width=1in,height=1.25in,clip,keepaspectratio]{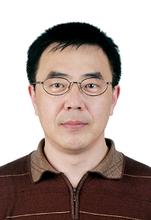}}]{Zhijian Wang}
	received the Ph.D. degree from Nanjing University, NanJing, China. He is currently a Professor with Hohai University, Nanjing, China. His current research interests include machine learning, computer vision and deep learning.
\end{IEEEbiography}
\begin{IEEEbiography}[{\includegraphics[width=1in,height=1.25in,clip,keepaspectratio]{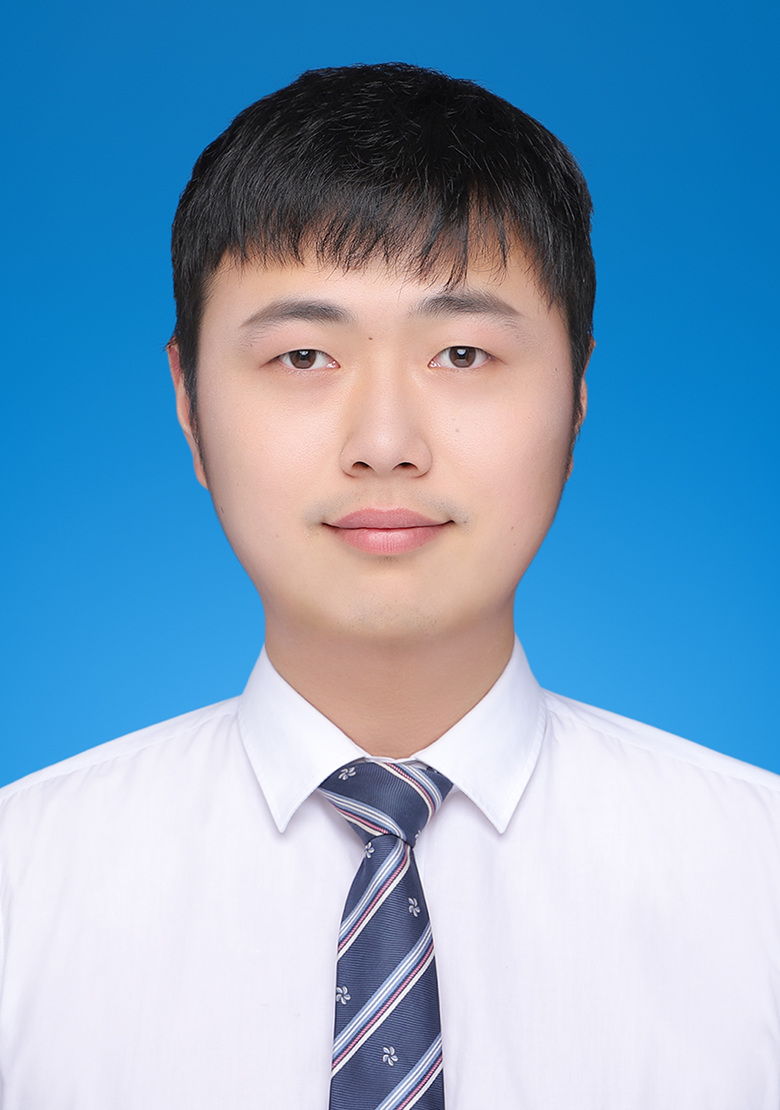}}]{Tianjin Yang}
	received the B.Sc. and M.Sc. degrees in Computer Science and Technology from Changzhou University, China, in 2018 and 2021, respectively. He is currently pursuing the Ph.D. degree with the Department of Artificial Intelligence, Hohai University, Nanjing, China. His current research interests include machine learning, computer vision.
\end{IEEEbiography}
\begin{IEEEbiography}[{\includegraphics[width=1in,height=1.25in,clip,keepaspectratio]{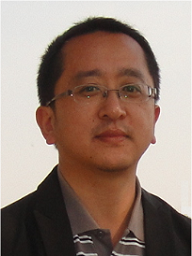}}]{Zhenjie Hou}
	received the Ph.D. degree in mechanical engineering from Inner Mongolia Agricultural University, in 2005. From 1998 to 2010, he was a Professor with the Computer Science Department, Inner Mongolia Agricultural University. In August 2010, he joined Changzhou University. His research interests include signal and image processing, pattern recognition, and computer vision.
\end{IEEEbiography}
\begin{IEEEbiography}[{\includegraphics[width=1in,height=1.25in,clip,keepaspectratio]{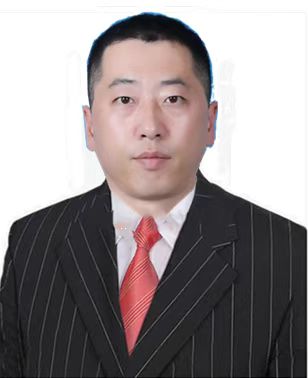}}]{Zhuang Miao}
	received the Ph.D. degree from PLA University of Science and Technology, Nanjing, China. Zhuang Miao is currently a professor of Army Engineering University of PLA, Nanjing, China. His current research focuses on artificial intelligence, pattern recognition and computer vision.
\end{IEEEbiography}
\end{document}